\documentclass[conference]{IEEEtran}
\IEEEoverridecommandlockouts
\usepackage{cite}
\usepackage{amsmath,amssymb,amsfonts}
\usepackage{algorithmic}
\usepackage{graphicx}
\usepackage{textcomp}
\usepackage{xcolor}
\def\BibTeX{{\rm B\kern-.05em{\sc i\kern-.025em b}\kern-.08em
    T\kern-.1667em\lower.7ex\hbox{E}\kern-.125emX}}
\begin{document}

\makeatletter
    \newcommand{\linebreakand}{%
      \end{@IEEEauthorhalign}
      \hfill\mbox{}\par
      \mbox{}\hfill\begin{@IEEEauthorhalign}
    }
    \makeatother

\title{Beyond-Labels: Advancing Open-Vocabulary Segmentation With Vision-Language Models\\

\thanks{
This work was supported by the Institute of Information \& Communications Technology Planning \& Evaluation (IITP) grant funded by the Korean government (MSIT) (No. RS-2024-00341055, Development of reinforcement learning-based automated driving AI software technology for optimal driving behavior decision in hazardous situations on congested roads.) \\

Accepted at the 17th IEEE International Conference on Advanced Computational Intelligence (ICACI 2025).

}
}


\author{\IEEEauthorblockN{Muhammad Atta ur Rahman}
\IEEEauthorblockA{\textit{Artificial Intelligence Creative Research Lab, ETRI} \\
\textit{University of Science and Technology, South Korea}\\
Daejeon, South Korea \\
rahman@etri.re.kr}
\and
\IEEEauthorblockN{Dooseop Choi}
\IEEEauthorblockA{\textit{Artificial Intelligence Creative Research Lab, ETRI} \\
\textit{University of Science and Technology, South Korea}\\
Daejeon, South Korea \\
d1024.choi@etri.re.kr}

\linebreakand
\IEEEauthorblockN{Seung-Ik Lee}
\IEEEauthorblockA{\textit{Field Robotics Research Section, ETRI} \\
\textit{University of Science and Technology, South Korea}\\
Daejeon, South Korea \\
the\_silee@etri.re.kr}
\and
\IEEEauthorblockN{KyoungWook Min}
\IEEEauthorblockA{\textit{Artificial Intelligence Creative Research Lab, ETRI} \\
Daejeon, South Korea \\
kwmin92@etri.re.kr}

}

\maketitle

\begin{abstract}
    Open-vocabulary semantic segmentation attempts to classify and outline objects in an image using arbitrary text labels, including those unseen during training. Self-supervised learning resolves numerous visual and linguistic processing problems when effectively trained. This study investigates simple yet efficient methods for adapting previously learned foundation models for open-vocabulary semantic segmentation tasks. Our research proposes ”Beyond-Labels,” a lightweight transformer-based fusion module that uses a handful of image segmentation data to fuse frozen visual representations with language concepts. This strategy allows the model to successfully actualize enormous knowledge from pre-trained models without requiring extensive retraining, making the model data-efficient and scalable. Furthermore, we efficiently capture positional information in images using Fourier embeddings, thus improving the generalization and resulting in smooth and consistent spatial encoding. We perform thorough ablation studies to investigate the major components of our proposed method in comparison to the standard benchmark PASCAL-5i, the method performs better despite being trained on frozen vision and language characteristics. 
\end{abstract}


\section{Introduction}
\label{sec:intro}

Recent research on segmentation has yielded promising results. Methods such as MoCo \cite{1} and DINO \cite{2} have demonstrated that powerful visual representations can be learned without utilizing labeled data. Interestingly, even though these models were only trained on unlabeled images, they appear to naturally detect objects in images, allowing for fundamental semantic segmentation \cite{3,4}.

A recent trend in segmentation is to combine vision and text domains. CLIP \cite{6}  and ALIGN \cite{7}, taught using simple contrastive learning techniques, have exhibited impressive capacities to perform new tasks without additional training, such as image classifying in a "zero-shot" way. As computational power increases, more powerful models are anticipated to emerge, trained on larger datasets and producing better image categorization results. It has opened a way towards the development of open-vocabulary segmentation, image retrieval, and multimodal reasoning within wide-ranging applications involving vision-language tasks. Larger and more powerful models are envisioned with increases in computational power and larger available datasets for further gains in both accuracy and robustness of image classification, along with other applications involving vision language. The above-mentioned continuous improvement illustrates the increased importance of contrastive learning in closing the gap in visual and textual understanding.

Semantic segmentation is a harder operation involving labeling each image pixel with a certain category. Recent models, such as FCN \cite{8}, U-Net \cite{9}, and DPT \cite{11}, have had tremendous success in training deep neural networks on huge datasets created for segmentation. Traditional techniques, however, rely on training classifiers for specific categories, which poses a significant constraint. These models can only predict labels for the categories on which they were taught; therefore, they cannot detect or segment objects from new, previously unseen categories during testing.

In our work, we investigate efficient techniques to connect the powerful pre-trained vision-language models to solve the challenge of open-vocabulary semantic segmentation successfully, "the segmentation of objects from any category based on their textual names or descriptions." To capture fine-grained spatial information, we first encode the input image using a frozen, self-supervised vision model supplemented with positional encoding. Fourier embedding is used to extract high-level spatial/frequency features from image data to help the model learn better. Simultaneously, we encode category names using a frozen language model to ensure a strong representation of textual descriptions. The integrated embeddings from two alternative modalities allow for exact matching of picture regions and textual descriptions, resulting in seamless segmentation across many categories. The key ingredient of our approach is the reliance on Fourier embeddings  \cite{24} to encode positional information without resorting to the commonly used pre-trained position embeddings. This enhances image generalization and results in smooth and consistent spatial encoding. A transformer-based \cite{38} fusion module receives the combined features. Taking into account data from the other modality, this module uses self-attention to update the features. To create the segmentation mask, we calculate the cosine similarity between each pixel and the text characteristics in each category. The dot product of the L2-normalized vision and language embeddings is used to accomplish this.\\

Among the contributions we provide, this includes a simple, lightweight, portable, transformer-based fusion module, Beyond-Labels, which stitches together effectively a series of open-vocabulary pre-trained models. To encode the position, unlike typical usage, common pre-trained position embeddings were not considered but instead are realized by Fourier embeddings, which bring great improvement in generalization among different images, but with smooth, steady spatial encoding guaranteed. This demonstrates strong zero-shot performance and competitive results on the PASCAL few-shot benchmark.


\begin{figure*}
    \centering
    \includegraphics[width=\textwidth]{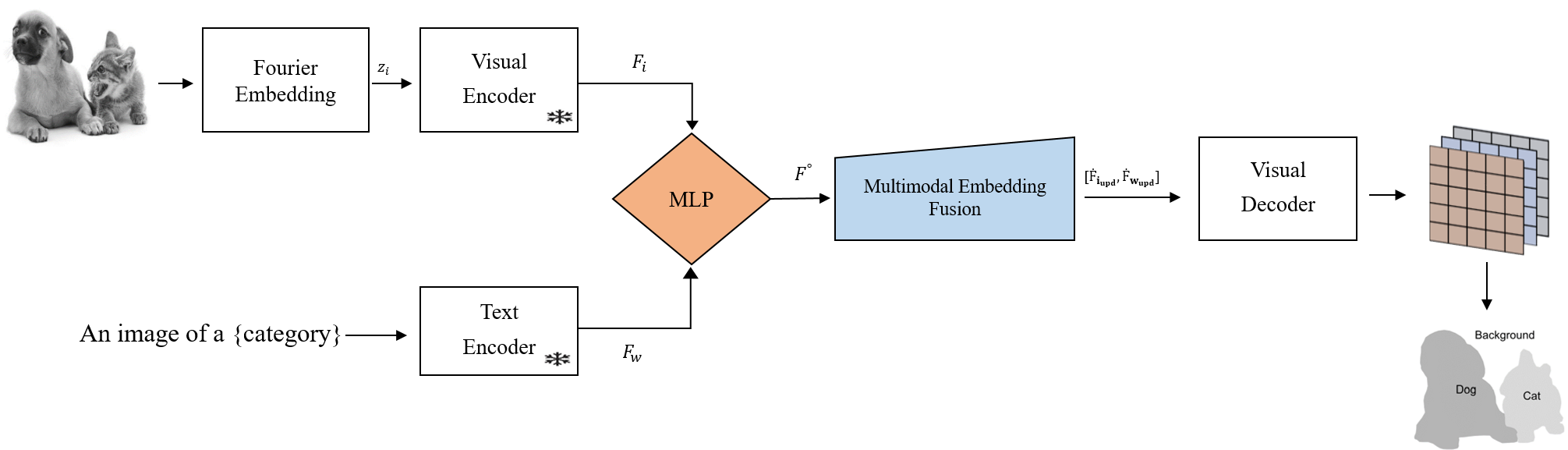}
    \caption{ Our proposed framework consists of a text encoder, a multimodal embedding fusion module, a decoder, and the vision encoder utilizing a pre-trained vision-language model with Fourier embeddings. The multimodal embedding fusion module combines the features that the frozen encoders have extracted from the text and image data. Before segmenting the visual features according to how closely they resemble text features, the decoder upscales them to their original size.}
    \label{fig:enter-label}
\end{figure*}
\section{Related Work}
\label{sec:formatting}

 We start by discussing semantic segmentation of novel classes, where word embeddings are used by zero-shot learning approaches to label unseen objects without training on extra data. Then, we talk about pre-trained language and vision models, wherein we present large-scale self-supervised models like CLIP that align images and text in the same vector space. Finally, we move on to Fourier embeddings for spatial encoding, which improves generalization by giving smooth, continuous positional representations for applications like image segmentation.
\subsection{Semantic Segmentation of Unseen Categories}

Teaching a network to segment new, previously unknown categories remains difficult, as most existing approaches only function with a predefined set of categories, requiring the test and training categories to match. Zero-shot semantic segmentation attempts to address this issue by employing word embeddings (semantic representations of category names) to recognize new categories without further training data. There are two primary approaches: Generative approaches, such as ZS3Net \cite{12} and CaGNet \cite{13}, which combine an image segmentation model with a mechanism for producing visual features for new categories via word embeddings, {Discriminative algorithms, such as SPNet \cite{14} map pixels and category names into a shared embedding space, which is then used to forecast the class of each pixel.

\subsection{Pre-trained Language and Vision Models}
Self-supervised learning has made substantial progress in computer vision and natural language processing. For example, methods such as contrastive learning (MoCo \cite{1}), metric learning (BYOL \cite{16}), and self-distillation (DINO \cite{2}) have emerged as effective tools for extracting characteristics for subsequent tasks. Language models, on the other hand, are commonly pre-trained with masked language modeling (BERT \cite{17}) or next-token prediction (GPT \cite{18}).

Large-scale vision-language models have recently received increased interest. CLIP \cite{6} is an important example, which trains on massive image-text pairs using contrastive learning and has successfully aligned images and text in a common space. Following that, CLIP's pre-trained knowledge is applied to various tasks, including object detection, picture captioning, image segmentation, and text-driven image manipulation. Instead of fine-tuning a pre-trained mode for each job, we adopt a simpler strategy, integrating a lightweight fusion module with pre-trained self-supervised vision and language models. This methodology is an incredibly effective foundation for the semantic segmentation of new groups.

 As foundation models \cite{19, 20} evolve rapidly, numerous studies have investigated how to merge multiple pre-trained models for different tasks. For example, Flamingo \cite{21} proposed an architecture that uses big, pre-trained vision-only and language-only models to perform image-language tasks with minimal data. Socratic Models \cite{22} uses a similar method but integrates pre-trained language models, vision-language models, and audio-language models to do multi-modal tasks like image captioning and video-to-text retrieval without retraining the models. Frozen \cite{23} suggested a method for aligning a pre-trained, frozen language model with a visual encoder using only a few examples and continual prompts. 

 \subsection{Fourier Embedding for Positional Encoding}

Recently, Fourier embeddings have become appealing alternatives to traditional position encodings in those tasks that require a strong spatial representation. Unlike learned positional embeddings, which will struggle hard to generalize outside the training distribution, Fourier-based embeddings provide a continuous and smooth encoding of spatial locations. Fourier embeddings turn out to be useful in improving generalization for implicit neural representations. For example, NeRF \cite{27} proposed Fourier feature mappings to help the positional encoding for 3D reconstruction tasks. Self-supervised learning for Perceiver IO \cite{28} uses Fourier embeddings to map high-dimensional inputs to a latent representation, with the best performance across various multimodal tasks. In semantic segmentation, Fourier network representations \cite{29} are utilized to encode better spatial dependencies, reducing large labeled datasets. Our approach thus replaces conventional learned embeddings with integrated Fourier embeddings of spatial information for smooth generalization across image domains. This goes in line with previous work showing that continuous and structured position encoding is helpful, both for vision and language models.

\section{Methodology}

This section begins by defining the open-vocabulary semantic segmentation problem. Next, we describe our suggested architecture, "Beyond-Labels," given in Fig 1, which links the previously trained language and vision models to accomplish the task.

\subsection{Problem Formulation}

 In this section, we will describe our approach to formulating the data. A training set is provided to us: \begin{align}
    \text{Data}_{\text{train}} &= \{(\mathcal{X}, \mathcal{Y}, \mathcal{U}) \mid \mathcal{X} \in \mathbb{R}^{H \times W \times 3}, \nonumber \\
    &\mathcal{Y} \in \mathbb{R}^{H \times W \times |\mathcal{U}|}, \nonumber \\
    &\mathcal{U} \subseteq \mathcal{S} \}
\end{align}

Where \( \mathcal{X} \) is the input picture, and \( \mathcal{Y} \) is the one-hot encoded segmentation mask for \( \mathcal{X} \); \( \mathcal{U} \) represents a unit of observed categories in \( \mathcal{X} \), and \( \mathcal{S} \) represents all training categories in this unit. We aim to train a segmentation model that separates a test image into significant areas of unexplored categories. 

\subsection{Visual Feature Extraction}

The process starts with splitting the input image \( \mathcal{X} \) as a collection of non-overlapping 2D patches. Each patch is flattened into a vector and put onto an embedding space with dimension 'd' using a linear projection.
Fourier embeddings \cite{24} are used to encode positional information rather than standard pre-trained positional embeddings, allowing for improved generalization to images while also maintaining smooth and consistent spatial encoding. Using sinusoidal functions, a f-emb is calculated for each patch at point (x, y) on the grid. The Fourier embedding is applied to the patch embedding, resulting in a final position-aware embedding for each patch.
\begin{align}
    X_i &= x_i + f\_emb(x, y)
\end{align}

Here \(x_i \) represents the flattened vector of the image patch. CLIP\cite{6}, a self-supervised language model, is used as a visual encoder.
\begin{align}
    F_i &={\textit{Vis\_Enc}}\left((X_i)\right) \in \mathbb{R}^{\mathcal{HW} \times d}
\end{align}

Here \textit{'d'} is the visual embedding dimension, \( \mathcal{H} = \frac{\mathcal{H}}{p} \) and \( \mathcal{W} = \frac{\mathcal{W}}{p} \) where \( p \) is the patch size.

\subsection{Textual Feature Extraction}

Semantic categories (\(W\)) are encoded using a text encoder to produce text embeddings.
\begin{align}
    F_W &={\textit{Text\_Enc}}(W) \in \mathbb{R}^{|W| \times d}
\end{align}

Where \textit{'d'} is the dimension of the word embeddings and \textit{'W'} is the number of input categories. We employ prompt templates such as "a picture of a category within the context" to provide semantic category context. The output embeddings generated by the text encoder are then averaged. The "Appendix Section" offers a collection of prompt templates. We use CLIP \cite{6}, a self-supervised language model, as a text encoder.
\begin{figure}
    \centering
    \includegraphics[width=1\linewidth]{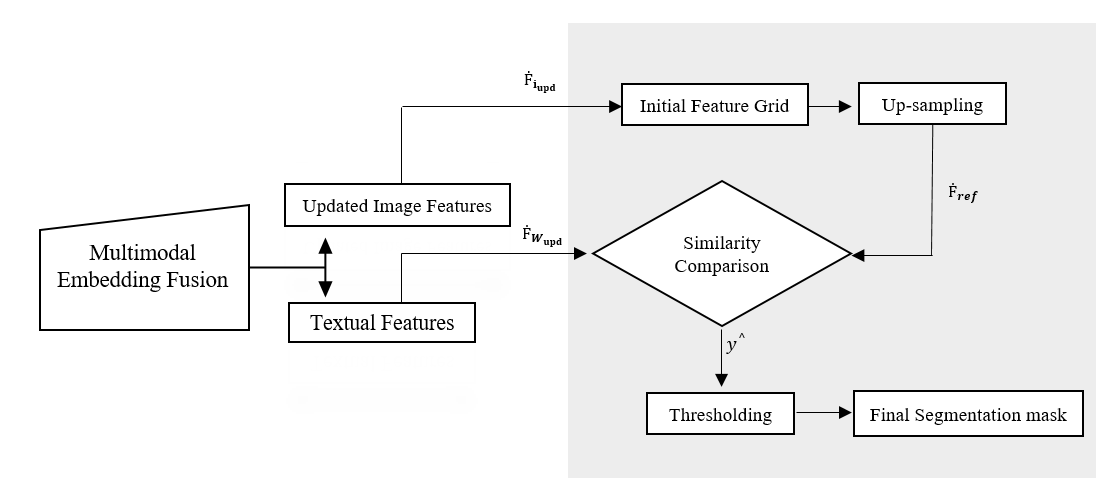}
    \caption{Overview of the decoder section. Multimodal embedding fusion combines visual and text encoder features that are processed through a decoder to generate upsampled visual features. These are then compared with the textual features using cosine similarity, creating a segmentation mask with labels refined by thresholding and a temperature-scaled sigmoid.}
    \label{fig:enter-label-2}
\end{figure}
\subsection{MultiModal Embedding Fusion}

To align the visual and textual information, we use MLPs (\( \theta\)) to align their channel dimensions (\( F_i \) and \( F_W \)). To capture the interactions between the textual and visual data, we subsequently pass the combined features through a multimodal embedding fusion module:

\begin{equation}
[F_{i\_upd}, F_{w\_upd}] = \textit{Fuse}(\theta [F_i, F_w])
\end{equation}
 The fusion module uses multiple transformer encoder layers to capture visual-text relationships. The fused visual \(F_{i\_upd}\) and text features \(F_{w\_upd}\) resemble the originals (\( F_i \)) and (\( F_w \)) in shape, but they are improved and enhanced by constantly focusing on one another's modality. This enrichment comes because the fusion module iteratively puts more focus on cross-modal interactions, which allow the features of the visual and text modalities to enrich each other's representations. Thus, the fused features become better aligned and may provide a more holistic representation of the underlying data, boosting performance in further tasks involving visuals or text. Such an attention-driven fusion process significantly enhances the model's capability of capturing fine-grained correlations between the two modalities, which is crucial for achieving high-quality results on tasks like open-vocabulary semantic segmentation. Our approach exploits the complementary strengths of both visual and text data to allow more accurate and context-aware embeddings

\subsection{The Hierarchical Visual Decoder}

The multimodal embedding fusion combines the visual and text encoders' features. These combined features are passed to the Hierarchical Visual Decoder for further processing. The decoder gives us the visual feature map, which is progressively upsampled to match the resolution of the input image. The upsampled visual features are compared with the textual features using cosine similarity to generate the logits \text{ŷ}. This comparison helps determine how closely each area in the image aligns with the provided text categories. The final result is a segmentation mask, where different regions of the image are assigned labels. Using class-wise thresholding and a sigmoid with a temperature (\( \mathcal{\tau} \)), final predictions are made using binary segmentation for each category.

\section{Experimental Results}

Our experiment section covers the PASCAL-5i dataset for few-shot segmentation evaluation, the evaluation Metric, the performance comparison across models, and the contribution of the different components of the proposed model. \\

\begin{table}[ht]
\caption{Performance comparison of different methods on the evaluation dataset of PASCAL-5\textsuperscript{i}.}
\centering  
\begin{tabular}{|c|c|c|c|c|c|c|}
\hline
\textbf{Model} & \textbf{Method} & \textbf{5\textsuperscript{0}} & \textbf{5\textsuperscript{1}} & \textbf{5\textsuperscript{2}} & \textbf{5\textsuperscript{3}}  & \textbf{mIoU} \\ \hline
LogReg  \cite{26}        & 1-shot & 26.9 & 42.9 & 37.1 & 18.4 & 31.4 \\ \hline
Siamese  \cite{26}       & 1-shot & 28.1 & 39.9 & 31.8 & 25.8 & 31.4 \\ \hline
Finetuning  \cite{26}    & 1-shot & 24.9 & 38.8 & 36.5 & 30.1 & 32.6 \\ \hline
1-NN  \cite{26}          & 1-shot & 25.3 & 44.9 & 41.7 & 18.4 & 32.6 \\ \hline
OSLSM \cite{26}          & 1-shot & 33.6 & 55.3 & 40.9 & 33.5 & 40.8 \\ \hline
CO-FCN  \cite{32}        & 1-shot & 36.7 & 50.6 & 44.9 & 32.4 & 41.1 \\ \hline
AMP-2  \cite{33}         & 1-shot & 41.9 & 50.2 & 46.7 & 34.7 & 43.4 \\ \hline
SG-One  \cite{34}        & 1-shot & 40.2 & 58.4 & 48.4 & 38.4 & 46.3 \\ \hline
PANet  \cite{35}         & 1-shot & 44.0 & 57.5 & 50.8 & 44.0 & 49.1 \\ \hline
SPNet \cite{14}          & zero-shot & 23.8 & 17.0 & 14.1 & 18.3 & 18.3 \\ \hline
ZS3Net \cite{12}         & zero-shot & 40.8 & 39.4 & 39.3 & 33.6 & 38.3 \\ \hline
\textbf{Ours}            & zero-shot & \textbf{46.6} & 56.2 & \textbf{51.6} & 43.2 & \textbf{49.4} \\ \hline
\end{tabular}
\label{tab:pascal_comparison}
\end{table}

\vspace{-10pt}  %

\begin{table*}[ht]
\centering
\caption{
Table representing object categories for different values of PASCAL-5\textsuperscript{i}.}
\renewcommand{\arraystretch}{1.5}
\setlength{\tabcolsep}{8pt}
\begin{tabular}{|c|c|c|c|}
    \hline
    \textbf{\textit{i} = 0} & \textbf{\textit{i} = 1} & \textbf{\textit{i} = 2} & \textbf{\textit{i} = 3} \\
    \hline
    \begin{minipage}{3.5cm}\centering
    aeroplane, bicycle, bird, boat, bottle
    \end{minipage} & 
    \begin{minipage}{3.5cm}\centering
    bus, car, cat, chair, cow
    \end{minipage} & 
    \begin{minipage}{3.5cm}\centering
    diningtable, dog, horse, motor-bike, person
    \end{minipage} & 
    \begin{minipage}{3.5cm}\centering
    potted plant, sheep, sofa, train, tv/monitor
    \end{minipage} \\
    \hline
\end{tabular}
\label{tab:categories}
\end{table*}

\begin{table*}[ht]
\caption{Ablation study on PASCAL-5\textsuperscript{i}. The \textbf{bold} indicates the best results.}
\centering
\renewcommand{\arraystretch}{1.5}
\setlength{\tabcolsep}{8pt} 
\begin{tabular}{|c|c|c|c|c|c|c|c|c|}
\hline
\textbf{Model} & \textbf{F\_Emb} & \textbf{Fusion} & \textbf{Visual\_Decoder} & \textbf{5\textsuperscript{0}} & \textbf{5\textsuperscript{1}} & \textbf{5\textsuperscript{2}} & \textbf{5\textsuperscript{3}} & \textbf{mIoU} \\ 
\hline
B\_L\_0 & \textbf{X} & \textbf{X} & \checkmark & 36.6 & 43.1 & 35.2 & 27.0 & 34.0 \\ 
\hline
B\_L\_1 & \checkmark & \textbf{X} & \checkmark & 40.4 & 46.7 & 38.4 & 32.3 & 39.5 \\ 
\hline
\textbf{B\_L\_2} & \checkmark & \checkmark & \checkmark & \textbf{46.6} & \textbf{56.2} & \textbf{51.6} & \textbf{43.2} & \textbf{49.4} \\ 
\hline
\end{tabular}
\label{tab:ablation}  
\end{table*}

\subsection{Dataset}

We evaluate our approach on the PASCAL-5\textsuperscript{i} \cite{26} dataset, which is designed for few-shot segmentation approaches. It is based on the well-known PASCAL VOC 2012 \cite{36} and SDB \cite{37} datasets. The dataset is divided into four folds (i = 0, 1, 2, 3) and contains 20 different object classes. Each fold contained five classes, and the model is trained on the other fifteen classes. This cross-validation design separates the classes used for testing from the classes used for training, ensuring that this is indeed a valid evaluation strategy to determine how well few-shot learning algorithms generalize to new categories. The support set contains some labeled instances, for example, 1-shot or 5-shot, from each of the five target classes in each fold to use as a reference for segmentation. Given this limited information in the support set, the query set should separate the objects precisely in the unlabeled photos that contain objects from the same five classes. Models trained on 15 classes and tested on the remaining 5 are doing so to cover all classes iteratively, using a technique called "leave-one-fold-out." Since this is indeed an exhaustive evaluation framework, it has made PASCAL-5i a standard benchmark for the comparison of few-shot segmentation. We have reported the results of our method in the ablation study and the results comparison section.

\subsection{Evaluation Metric}
The key evaluation metric is mean intersection-over-union (mIoU), which averages IoU across all classes within the fold. 
\[
\text{mIoU} = \frac{1}{F} \sum_{F=1}^{F} \text{IoU}_F
\]
Where \text{IoU} denotes the intersection over the union of class (\text{F}),  which is the target fold's class count.

\subsection{Results Comparison}

In terms of mIoU, Table 1 indicates a comparison of different models of segmentation on all folds. The worst scores belonged to Folds 2 and 3, indicating the incapability of SPNet to go against the native complexities of classes in few-shot segmentation tasks based on generalization power, where the overall mIoU reached 18.3. ZS3Net outperformed SPNet with a score of 38.3 on mIoU for generalizing semantic features across folds. OSLSM outperforms ZS3Net and SPNet with an mIoU of 40.8; however, there are problems with some folds, such as fold 3. OSLSM also compared their method's performance against a range of baselines, including LogReg, Siamese, Finetuning, and 1-NN, and adopted previous work on dense pixel prediction as baselines against which they compare. PANet and CO-FCN are both competitive, with PANet having an mIoU of 49.1 and CO-FCN having an mIoU of 41.1, outperforming OSLSM and ZS3Net. Improvements are also seen in AMP-2 and SG-One, with AMP-2 obtaining an mIoU of 43.4 and SG-One getting 46.3 and outperforming in multiple folds (folds 1 and 2). With an mIoU of 49.4, our model outperforms all baselines; it performs best at folds 0 and 2, respectively. This again supports the assumption that integrating Fourier embedding with multimodal fusion, along with a visual decoder, improved feature representation by providing maximum consideration to the reciprocal textual and visual information. This indicates that the increase of 8.6 in mIoU compared to OSLSM, 5.3 compared to PANet, and 11.1 compared to ZS3Net demonstrates the effectiveness of our proposed components in handling complex spatial relationships and unseen class distributions. These results confirm that not only does our approach achieve more precise segmentation, but it also achieves more balanced performance across folds.

\subsection{Ablation Study}

 We investigated the contribution of each core component in the proposed "Beyond-Labels" framework. Specifically, we investigated the impact of the Fourier Embedding (F\_Emb), multimodal embedding fusion module (Fusion), and the use of a decoder instead of traditional positional embedding techniques. Detailed results of this study can be seen in Table 3 for breaking down performance in different configurations.
 
\begin{itemize}
    \item  By default, it uses only an upward model that includes the usual positional embedding and upsampling using a decoder. Using this standard baseline model resulted in minimum performance, by which B\_L\_0 obtained an mIoU of 34.0. The large decreases in scores in fold 3 represent a limitation based on the use of standard position embedding without robust feature fusion in action. Without Fourier embedding, the model cannot capture high-dimensional and expressive positional information, critical for improving segmentation accuracy, especially in challenging folds such as folds 2 and 3.
    \item  While for B\_L\_1, incorporation of Fourier embedding improves its performance on all counts, with the mIoU increasing to 39.5. This, in turn, indicates that adding frequency-based representation is vital for boosting the feature extraction required to improve the generalization of non-visible classes. More important is the model's adaptation capability to different complicated spatial scenarios due to improved performance on all folds.
    \item  The best-performing model, B\_L\_2, is the full model, including the Fourier embedding, fusion, and decoder, with an mIoU of 49.4. Additional positional information is introduced by Fourier Embedding that, combined with the fusion module, enables the model to learn richer representations. Further refinement of the upsampling process with the decoder leads to consistent improvements in all the folds. This includes the Fourier embedding module with the Fusion module, which generates high-dimensional and expressive forms of position-based information, thus yielding significant gains in segmentation accuracy. This further facilitates better spatial understanding within the image, boosting the capability for more precise feature representations of the model. \\
\end{itemize}

The benefits of Fourier embedding are easily found in all folds in B\_L\_1 and B\_L\_2, where traditional embeddings often fall short.

\section{Conclusion}
Our research aims to make advanced vision and language models more helpful for open-vocabulary semantic
segmentation. We introduce "Beyond-Labels," a compact and efficient module that links vision and language models to do open-vocabulary semantic segmentation, which involves finding and categorizing items in images, even for
previously unknown categories. We tested it on the “PASCAL” dataset and found that it performs extremely well,
exceeding some existing approaches. Despite its simplicity, “Beyond-Labels” delivers excellent results. To assess our work’s resilience and enhance generalization across many domains, we will expand it to include more datasets and perform more ablation studies. Through testing on various datasets, we hope to ensure that our method works reliably and successfully in practical situations, capturing a wider range of difficulties and subtleties in semantic segmentation tasks.


\section{Appendix}
\subsection{Template List}
\begin{itemize}
    \item An image of a \{category\}.
    \item This is an image of a \{category\}.
    \item An image of a small \{category\}.
    \item An image of a medium \{category\}.
    \item An image of a large \{category\}.
    \item An image of a \{category\} within the context.
    \item An image of the \{category\} within the context.
    \item An image of the \{category\} within the context.
    \item A resized image of a\{category\} within the context.
    \item This falls under a \{category\} within the context.
    \item This falls under the \{category\} within the context.
    \item This falls under one \{category\} within the context.
\end{itemize}

{\small
\bibliographystyle{ieee_fullname}
\bibliography{egbib}
}

\end{document}